%% file: main.tex
\documentclass{article}
\usepackage{spconf,amsmath,graphicx}
\usepackage{booktabs}
\usepackage{xcolor}
\usepackage{amssymb}
\usepackage{subfigure}
\usepackage{svg}
\usepackage{bbm}
\usepackage{multirow}

\title{Rethink long-tailed recognition with Vision transformers}
\name{Zhengzhuo Xu$^{1 \ast}$, Shuo Yang$^{1 \ast}$, Xingjun Wang$^1$, Chun Yuan$^{1,2 \dagger}$\thanks{$\ast$: Equal Contribution, \ \ $\dagger$: Corresponding Author. This work was supported by the National Key R\&D Program of China (2022YFB4701400/4701 402), SZSTC Grant(JCYJ 20190809172201639, WDZC2020082020065500 1), Shenzhen Key Laboratory (ZDSYS20210623092001004).}}
\address{$^1$Tsinghua Shenzhen International Graduate School, Tsinghua University, $^2$Shenzhen Peng Cheng Lab}

\begin{document}
\maketitle
\begin{abstract}
    In the real world, data tends to follow long-tailed distributions w.r.t. class or attribution, motivating the challenging Long-Tailed Recognition (LTR) problem. In this paper, we revisit recent LTR methods with promising Vision Transformers (ViT). We figure out that 1) ViT is hard to train with long-tailed data. 2) ViT learns generalized features in an unsupervised manner, like mask generative training, either on long-tailed or balanced datasets. Hence, we propose to adopt unsupervised learning to utilize long-tailed data. Furthermore, we propose the Predictive Distribution Calibration (PDC) as a novel metric for LTR, where the model tends to simply classify inputs into common classes. Our PDC can measure the model calibration of predictive preferences quantitatively. On this basis, we find many LTR approaches alleviate it slightly, despite the accuracy improvement. Extensive experiments on benchmark datasets validate that PDC reflects the model's predictive preference precisely, which is consistent with the visualization.
\end{abstract}
\begin{keywords}
metric, long-tailed learning, vision transformers, representation learning, imbalanced data.
\end{keywords}

\input{Texs/1.intro.tex}
\input{Texs/3.method.tex}

\input{Texs/4.exp.tex}
\input{Texs/5.conclusion.tex}

\clearpage

{\small
\bibliographystyle{IEEEbib}
\bibliography{main}
}
\clearpage

\end{document}

%% file: Texs/1.intro.tex
\section{Introduction}
\label{sec:intro}
With rapid advances in visual classification, deep models tend to depend on balanced large-scale datasets more seriously \cite{Imagenet, COCO}. However, the number of instances in real-world data usually follows a Long-Tailed (LT) distribution w.r.t. class. Many tail classes are associated with limited samples, while a few head categories occupy most of the instances \cite{CB-loss, OLTR, PriorLT, icasspunbalance}. The model supervised by long-tailed data tends to bias toward the head classes and ignore the tail ones. The tail data paucity makes the model hard to train with satisfying generalization. It is still a challenging task to overcome Long Tailed Recognition (LTR) and utilize real-world data effectively.

Recent literature mainly adopt two approaches to tackle LT data, i.e., feature re-sampling and class-wise re-weighting. The re-sampling methods balanced select the training data by over-sampling the tail or under-sampling the head. Some effective proposals replenish the tail samples via generation or optimization with the help of head instances\cite{M2m, Bagoftricks}. The re-weighting ones punish different categories with data number relevant weight or logit bias\cite{BalCE, LADE-loss}. Although the aforementioned methods have greatly mitigated the LT problem, the conclusions hold on the ResNet-based backbones \cite{MiSLAS, IB-loss}.

In recent years, many transformer-based backbones \cite{vit} have surpassed the performance of CNN. DeiT\cite{deit3} proposes an effective receipt to train ViT with limited data, and MAE\cite{MAE} adopts a masked autoencoder to pre-train the ViT. However, there is limited research on how ViTs perform on LTR. Motivated by this, we rethink the previous LT works with ViT. We figure out that it is hard to train ViTs with long-tailed data while the unsupervised pretraining manner ameliorates it by a large margin. The unsupervised pretraining ViTs will learn meaningful feature (c.f. Figure \ref{fig:mae-pretrain}) and generalize well on downstream tasks (c.f. Table \ref{tab:recipe}), either on long-tailed or balanced datasets.

Numerous studies have demonstrated that the model supervised by the LT dataset will inevitably exhibit prediction bias to the head\cite{PriorLT, LA,  LADE, icassphead}. The predictor will simply classify the inquiry image to the head to attain a low misclassification error. The previous metrics, like accuracy on the validation dataset, are difficult to evaluate the model's predictive preference directly. The same accuracy may come at the cost of a different number of predictions (c.f. Figure \ref{fig:teaser}). Although some works show models' prediction distribution by visualization qualitatively\cite{PriorLT, MiSLAS, NCL}, a metric is required to evaluate it quantitatively. In this paper, we propose Prediction Distribution Calibration (PDC) to fill this gap. Specifically, if we view the prediction number and target instance number of each class as probability distributions, we can measure the distance between the two probability distributions. Considering the imbalance degree of training samples, we take the training label into account as well. To summarize, our main contributions are:

\noindent \textbf{1)} We figure out that it is difficult to train ViT with long-tailed data, which can be tackled with unsupervised pretraining.

\noindent \textbf{2)}  We propose PDC to provide a quantitative view to measure how the proposal ameliorates the model predictive preference.

\noindent \textbf{3)}  We conduct extensive experiments to analyze LTR proposals' performance on ViT with our proposed PDC, which will accurately indicate the model's predictive bias and is consistent with the visualization results.

\input{Figs/teaser.tex}

%% file: Figs/teaser.tex
\begin{figure*}[!t]
\centering
\includegraphics[width=1\linewidth]{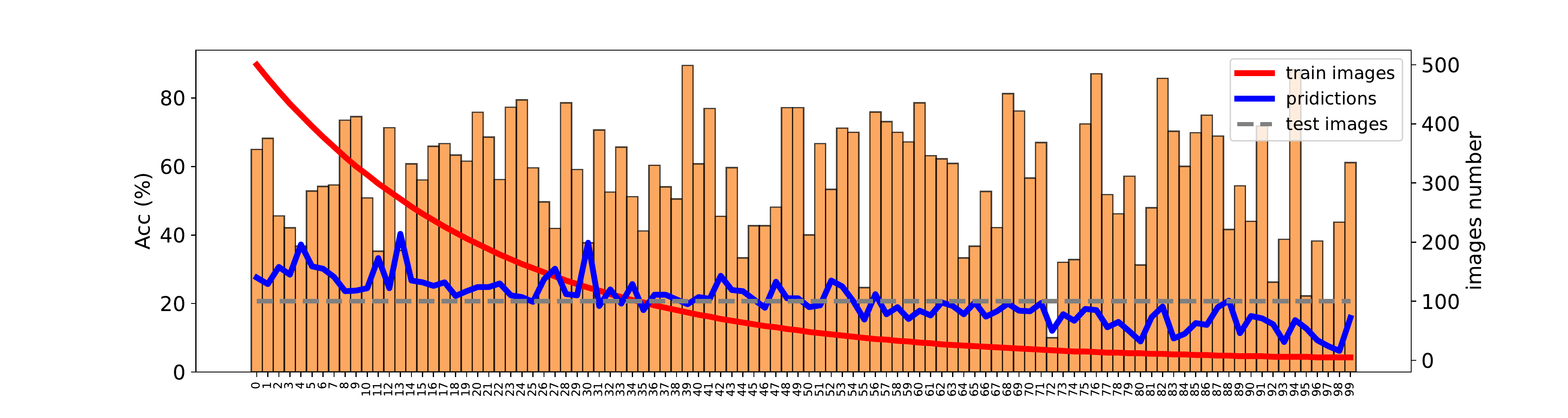}
\vspace{-20pt}
\caption{Visualization of ViT-B on CIFAT100-LT (IF=100). \textbf{Acc trap}: the accuracy can not reflect predictive bias. A class can obtain on par accuracy with another one with much more predictions. Take class 0 and 99 for an illustration.}
\label{fig:teaser}
\vspace{-10pt}
\end{figure*}

%% file: Texs/3.method.tex
\section{The proposed approach}
\subsection{Long Tail Recognition}
\label{LTR}
Given an $C$-classes labeled dataset containing $N$ training instances, $\mathbf{D}=\left\{\left(x_1, y_1\right),\left(x_2, y_2\right), \ldots,\left(x_n, y_n\right)\right\}$, where $y_i \in \mathcal{C} = \{1,...,C\}$ and the distribution is $\mathbb{P}(\mathbf{x},\mathbf{y})$. 
In this paper, we define a base classification model as ${\mathcal{M}_\theta}$, which is parameterized by $\theta$. For each input image $x$, the output logits as ${\mathbf{z}_\theta(x)=\mathcal{M}(x|\theta) }=\{z_1,...,z_c\}$. The goal is to optimize the parameters $\theta$ to get the best estimation of $\mathbb{P}(\mathbf{x},\mathbf{y})$. Generally, one adopts \textit{softmax} function to map the output ${\mathcal{M}_(x|\theta)}$ as the conditional probability:
\begin{equation}
p\left(\mathbf{y} \mid \mathbf{x}; \theta\right)=\frac{e^{\mathcal{M}(\mathbf{x} \mid \theta)_{\mathbf{y}}}}{\sum_i e^{\mathcal{M}(\mathbf{x} \mid \theta, \phi)_{\mathbf{y}_i}}}
\end{equation}

We get the posterior estimates $\mathbb{P}(y|x):=p\left(\mathbf{y}|\mathbf{x}; \theta\right)$ by maximum likelihood $\mathbb{P}(x|y)$ estimation, which is represented by model parameters $\theta$. In LTR, we train the model with long-tailed distributed training data $\mathbb{P}_{s}(x,y)$ while evaluating it with uniform ones $\mathbb{P}_{t}(x,y)$. The label prior distribution $\mathbb{P}_s(y)$ will be different for each class while keeping consistent in the test dataset, i.e., $\mathbb{P}_t(y):=1/C$. For a tail class $i$, $\mathbb{P}_s(y_i) \ll \mathbb{P}_t(y_i)$. 
According to the Bayesian Theory, the posterior is proportional to the prior times the likelihood. Considering the same likelihood, i.e., $\mathbb{P}_s(x|y) = \mathbb{P}_t(x|y)$, we have the posterior on the target dataset:
\begin{equation}
\mathbb{P}_t(y_i| x)=\mathbb{P}(x|y_i) \cdot \mathbb{P}_s(y_i) / \mathbb{P}_t(y_i)
\label{bayes}
\end{equation}

With the Eq.\ref{bayes} and balanced target  distribution $\mathbb{P}_t(y):=1/C$, we have $\mathbb{P}_t(y_i| x) \propto \mathbb{P}_s(y_i)$. Therefore, models tend to \textit{predict a query image into head classes} to satisfy the train label distribution $\mathbb{P}_s(y_i)$, which is called \textbf{predictive bias}. Such a mismatch makes the generalization in LTR extremely challenging, and the traditional metrics, e.g., mean accuracy on the training dataset, exacerbates biased estimation when evaluating models on the balanced test set.

\subsection{Vision Transforms}
ViT reshapes a image $x\in \mathbb{R}^{H \times W \times C}$ into a sequence (length $L=H\times W / P^2$) of flattened 2D patches $x_P \in\mathbb{R}^{L \times (P^2 \cdot C)}$, where $H \times W$ are the resolution of $x$, $C$ is channels, $P$ is the patch resolution. Although ViTs perform well on numerous visual tasks, we figure out that \textit{it is hard to train ViTs with long-tailed data, and the performance is unsatisfactory}. Recent work trains ViTs without label supervision by the encoder ($\mathcal{E}$) decoder ($\mathcal{D}$) architecture and random mask $\mathbf{M}$:
\begin{equation}
\hat{\mathbf{x}}=\mathcal{D}\left(\mathcal{E}(\mathbf{M} \odot \mathbf{x})\right)
\label{mae}
\end{equation}
\vspace{-10pt}

\input{Tabs/Cifar.tex}

We pinpoint that the \textit{ViTs will learn generalized feature extraction by Eq.\ref{mae}, either on long-tailed or balanced datasets}. Such an observation inspires us to adopt it as a strong baseline to evaluate the performance with ViTs.

\subsection{Predictive Distribution Calibration}
In LTR, recent works try to compensate for the mismatch of $\mathbb{P}_s(y)$ and $\mathbb{P}_t(y)$, which is described in section \ref{LTR}. However, they all adopt the Top1-accuracy to evaluate their proposals, which fails to show whether the mismatch is fixed. To fill the gap and measure it intuitively, we propose the Predictive Distribution Calibration (PDC) to quantitative analyze the model's predictive bias.

\input{Figs/mae_pretrain.tex}

\noindent \textbf{Step 1}: Here, we view the prediction number w.r.t. class as the predictive distribution $\hat{\mathbb{P}}_{t}(y)$. Considering the balanced label distribution $\mathbb{P}_{t}(y)$, we can calculate the \textit{distance} between the above two distributions. Considering to measure this \textit{distance} via Kullback-Leibler divergence (KL), we have:
\begin{equation}
    D(\mathbb{P}_t, \hat{\mathbb{P}}_t) = \frac{1}{C} \sum_{y_i\in \mathcal{C}} \mathbb{P}_{t}(y_i) \cdot \left[\log \mathbb{P}_{t}(y_i) - \log \hat{\mathbb{P}}_{t}(y_i)\right] \!
\end{equation}

\noindent \textbf{Step 2}: Generally, the larger gap between $\mathbb{P}_{s}(y)$ and $\mathbb{P}_{t}(y)$, the more difficult to overcome the model predictive bias. To eliminate it, we take the training label distribution $\mathbb{P}_{s}(y)$ into consideration, which can be written as $D(\mathbb{P}_t, \mathbb{P}_s)$:
\begin{equation}
\begin{aligned}
       &PDC(\mathcal{M}_\theta, \mathbf{D}) = D(\mathbb{P}_t, \hat{\mathbb{P}}_t) / D(\mathbb{P}_t, \mathbb{P}_s) \\
       &= \frac{\sum_{y_i\in \mathcal{C}} \mathbb{P}_{t}(y_i) \cdot \log \mathbb{P}_{t}(y_i) - \mathbb{P}_{t}(y_i) \cdot \log \hat{\mathbb{P}}_{t}(y_i)}{ \sum_{y_i\in \mathcal{C}}\mathbb{P}_{t}(y_i) \cdot \log \mathbb{P}_{t}(y_i) - \mathbb{P}_{t}(y_i) \cdot \log \mathbb{P}_{s}(y_i)} 
\end{aligned}
\label{D-kl}
\end{equation}

\noindent \textbf{Step 3}: Notice that $D(\mathbb{P}_t, \mathbb{P}_s)$ will be zero when the target label distribution is consistent with the training label distribution. Hence, we add an extra $\varepsilon=1e-6$ to $D(\mathbb{P}_t, \mathbb{P}_s)$ for numerical stability.

\subsection{Further Analysis}
Previous work evaluates the model predictive bias in the following manners:

\noindent \textbf{Group Acc} divides $\mathcal{C}$ into several groups $\{\mathcal{G}_1, \mathcal{G}_2,..., \mathcal{G}_n\}$ according to the $\mathbb{P}_s(y)$, where $\forall i, \mathcal{G}_i \subseteq \mathcal{C}$. A widely adopted group type is \{\textit{Many}, \textit{Medium}, \textit{Few}\} and the accuracy of each group can be calculated by:
\begin{equation}
Acc(\mathcal{G}) = \frac{1}{N_\mathcal{G}} \sum_{y \in \mathcal{G}} \mathbb{I}\left(y = \operatorname{argmax}_{y_{i} \in \mathcal{G}} \mathcal{M}(\mathbf{x}|\theta)_{y_{i}}\right),
\label{group-acc}
\end{equation}
where $N_\mathcal{G}$ is the sum instance number in $\mathcal{G}$ and $\mathbb{I}\left(\cdot\right)$ is indicator function. However, the weakness is obvious: 1) $Acc(\mathcal{G})$ heavily depends on $\mathbb{P}_s(y)$ and the definition of group $\mathcal{G}$. 2) The \textit{Few} accuracy can not avoid the acc trap (see Figure \ref{fig:teaser}).

\noindent \textbf{Confusion matrix} is used to visualize the classification situation for each class. However, 1) it can not quantitatively measure how much the predictive bias the methods alleviate. 2) It will be unintuitive when the class number $C$ gets larger. As a comparison, our PDC is plug-and-play with negligible computation operation. With fixed model structure and datasets, we can compare proposed methods quantitatively.

\input{Tabs/ViT.tex}

%% file: Tabs/Cifar.tex
\begin{table*}[t]
\centering
\caption{Performance on CIFAR100-LT. Acc@R: Top1 accuracy with ResNet32. Acc@V: Top1 accuracy with ViT-B.}
\begin{tabular}{l|cccccccccc}
\hline
\multicolumn{1}{c|}{Imbalance} & \multicolumn{3}{c|}{10}                   & \multicolumn{3}{c|}{50}                   & \multicolumn{3}{c|}{100}                  & \multirow{2}{*}{var.} \\ \cline{1-10}
\multicolumn{1}{c|}{Method}    & Acc@R   & Acc@V   & \multicolumn{1}{c|}{PDC$\downarrow$}  & Acc@R   & Acc@V   & \multicolumn{1}{c|}{PDC$\downarrow$}  & Acc@R    & Acc@V   & \multicolumn{1}{c|}{PDC$\downarrow$}  &                       \\ \hline
CE                             & 55.70 & 66.02 & \multicolumn{1}{c|}{0.34} & 43.90 & 54.78 & \multicolumn{1}{c|}{0.62} & 38.30 & 50.59 & \multicolumn{1}{c|}{0.64} & 0.03                  \\
BCE \cite{Deit}                           & -     & 64.63 & \multicolumn{1}{c|}{0.46} & -     & 50.65 & \multicolumn{1}{c|}{1.31} & -     & 45.50 & \multicolumn{1}{c|}{2.25} & 0.80                  \\
CB  \cite{CB-loss}                           & 57.99 & 66.30 & \multicolumn{1}{c|}{0.06} & 45.32 & 56.18 & \multicolumn{1}{c|}{0.08} & 45.32 & 50.63 & \multicolumn{1}{c|}{0.13} & 0.00                  \\
LDAM  \cite{LDAM-loss}                         & 56.91 & 63.99 & \multicolumn{1}{c|}{0.47} & 45.00 & 54.53 & \multicolumn{1}{c|}{0.61} & 39.60 & 50.39 & \multicolumn{1}{c|}{0.82} & 0.03                  \\
MiSLAS \cite{MiSLAS}                        & \textbf{63.20} & 66.65 & \multicolumn{1}{c|}{0.26} & \textbf{52.30} & 55.90 & \multicolumn{1}{c|}{0.56} & 47.00 & 50.62 & \multicolumn{1}{c|}{0.94} & 0.12                  \\
LADE \cite{LADE-loss}                          & 61.70 & \textbf{68.32} & \multicolumn{1}{c|}{0.07} & 50.50 & 60.03 & \multicolumn{1}{c|}{0.10} & 45.40 & \textbf{57.25} & \multicolumn{1}{c|}{0.10} & 0.00                  \\
IB \cite{IB-loss}                            & 57.13 & 65.12 & \multicolumn{1}{c|}{0.06} & 46.22 & 43.78 & \multicolumn{1}{c|}{1.09} & 42.14 & 42.30 & \multicolumn{1}{c|}{0.46} & 0.27                  \\ 
BalCE \cite{BalCE}                         & 63.00 & 68.11 & \multicolumn{1}{c|}{\textbf{0.04}} & 49.76     & \textbf{60.67} & \multicolumn{1}{c|}{\textbf{0.04}} & \textbf{50.80} & 56.86 & \multicolumn{1}{c|}{\textbf{0.05}} & \textbf{0.00}                  \\\hline
\end{tabular}
\vspace{-5pt}
\label{tab:cifar}
\end{table*}

%% file: Figs/mae_pretrain.tex
\begin{figure*}[!t]
\centering
    \subfigure[Input]{
    \includegraphics[height=0.14\linewidth]{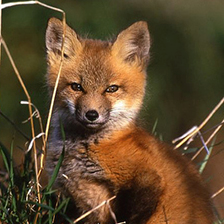} }
    \subfigure[Masked]{
    \includegraphics[height=0.14\linewidth]{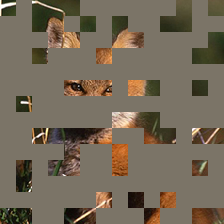} }
    \subfigure[Reco.+LT]{
    \includegraphics[height=0.14\linewidth]{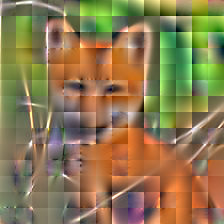} }
    \subfigure[Reco.+BAL]{
    \includegraphics[height=0.14\linewidth]{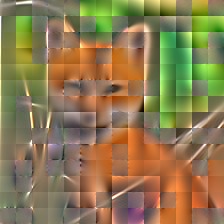} }
    \subfigure[Reco.+LT+U]{
    \includegraphics[height=0.14\linewidth]{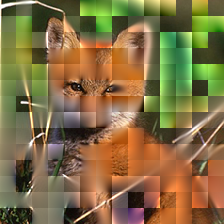} }
    \subfigure[Reco.+BAL+U]{
    \includegraphics[height=0.14\linewidth]{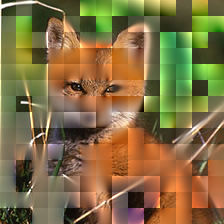} }
\vspace{-10pt}
\caption{Reconstruction visualization of MAE. LT: pretrain with long-tailed data. BAL: pretrain with balanced data. LT and BAL have the same total instances. U: add unmasked patch. ViTs pretrained on both LT and BAL learn meaningful features.}
\vspace{-10pt}
\label{fig:mae-pretrain}
\end{figure*}


%% file: Tabs/ViT.tex
\begin{table}[t!]
\centering
\vspace{-5pt}
\caption{Training recipe of ViT. IN-LT: the long-tailed subset of ImageNet. IN-BAL: the balanced subset of ImageNet.}
\begin{tabular}{ccccc}
\hline
Dataset   &Images   & ViT\cite{vit} & DeiT III\cite{deit3} & MAE\cite{MAE}   \\ \hline
IN-LT    & 18.6K  & 31.63   & 48.44    & 56.48 \\
IN-BAL   & 18.6K  & 38.66   & 65.24    & 69.11 \\ \hline
\end{tabular}
\vspace{-10pt}
\label{tab:recipe}
\end{table}

%% file: Texs/4.exp.tex
\section{Experiments}
\label{sec:exp}

\input{Figs/cf.tex}

\subsection{Datasets}
\textbf{CIFAR100-LT} \cite{cifar} is created from the original CIFAR datasets that have 100 classes with 60K images. The skewness of the dataset is controlled by an Imbalance Factor (IF), which is the ratio between the most and the least frequent classes. We follow previous work\cite{LDAM-loss, CB-loss} to utilize the dataset with $IF = [10, 50, 100]$ for comprehensive comparisons.

\noindent\textbf{iNaturalist 2018} \cite{inaturalist} is the large-scale real-world dataset for LTR with 437.5K images from 8,142 classes. It is extremely imbalanced, with an imbalance factor of 500. We use the official training and validation split in our experiments.

\subsection{Implement Details}
We use a pre-trained ViT-Base model from MAE and fine-tune it with $32$ (CIFAR-LT) and $128$ (iNat18) resolution. We use AdamW optimizer with momentum $\beta_1 = 0.9$ and  $\beta_2 = 0.999$. We train the model for 100 epochs with an effective batch size 1024 and weight decay 0.1. The base learning rate is $1e-3$, which follows cosine decay with 5 warmup epochs. We use Mixup (0.8) and Cutmix (1.0) as augmentation and set the drop path of ViT to 0.1.

\subsection{Compared Methods}
We adopt the recipe in vanilla ViT\cite{vit}, DeiT III\cite{deit3}, and MAE\cite{MAE} to train ViTs. In view of MAE's excellent performance and low computation consumption, we adopt MAE for our following evaluation. We adopt vanilla CE loss, Binary CE\cite{deit3}, BalCE\cite{BalCE}, CB\cite{CB}, LDAM\cite{LDAM-loss}, MiSLAS\cite{MiSLAS}, LADE\cite{LADE}, and IB loss\cite{IB-loss} for comprehensive comparisons. We ignore the multi-expert (heavy GPU memory) and contrastive learning (contradictory to MAE) methods.

\noindent Table \ref{tab:recipe} shows the results of different training manners. With the same training image number, the LT is lower than BAL for all recipes. MAE achieves the best on two datasets and learns meaningful features on both datasets (Figure \ref{fig:mae-pretrain}). Hence, we select MAE for the following experiments.

\input{Tabs/iNat18.tex}

\subsection{LTR Performance with ViT}
It is challenging to train ViTs directly on LTR datasets (Table \ref{Tab:apdx}), because it is difficult to learn the inductive bias of ViTs and statistical bias of LTR (Eq.\ref{bayes}) simultaneously (Table \ref{tab:recipe}). In Table \ref{tab:cifar}, we mainly re-rank different losses of LTR on ViT-Base, which is based on the pre-trained weights on ImageNet. The results in Table \ref{tab:inat} are trained \textit{from scratch} in the MAE manner without pre-trained weights to show the performance gap between ResNet and ViT. We only conduct the architecture comparisons on the iNat18 because ViTs are hard to train from scratch with limited data and resolution, like CIFAR. 

As Table \ref{tab:cifar} \& \ref{tab:inat} show, BalCE achieves satisfying performance on both datasets, which indicates its effectiveness and generalization. Compared to the performance on ResNet, MiSLAS shows poor Acc and PDC, which means its special design is hard to generalize on ViT. In addition, IB is difficult to train for its numerical instability and thus results in worse performance (7\%$\downarrow$). For most proposals, the performance of PDC keeps consistent with Top-1 Acc and Few Acc. However, LDAM has better accuracy and worse PDC compared to CB, which means it alleviates predictive bias slightly. We additionally calculate the variance of PDC for the different unbalanced degrees, as shown in Table \ref{tab:cifar}. From this point of view, BCE obtains the maximum variance with decreasing performance, which suggests its weak adaptability.

Figure \ref{fig:cf} presents the visualization with confusion matrix. A larger PDC indicates more centralized off-diagonal elements(e.g., BCE). BalCE makes more balanced predictions with a smaller PDC, which demonstrates PDC is a precise quantitative metric to measure prediction distributions.

\input{Tabs/Rebuttal}

%% file: Figs/cf.tex
\begin{figure*}[t]
\centering
\resizebox{1\textwidth}{30mm}{
    \subfigure[BCE]{
    \includegraphics[height=0.22\linewidth]{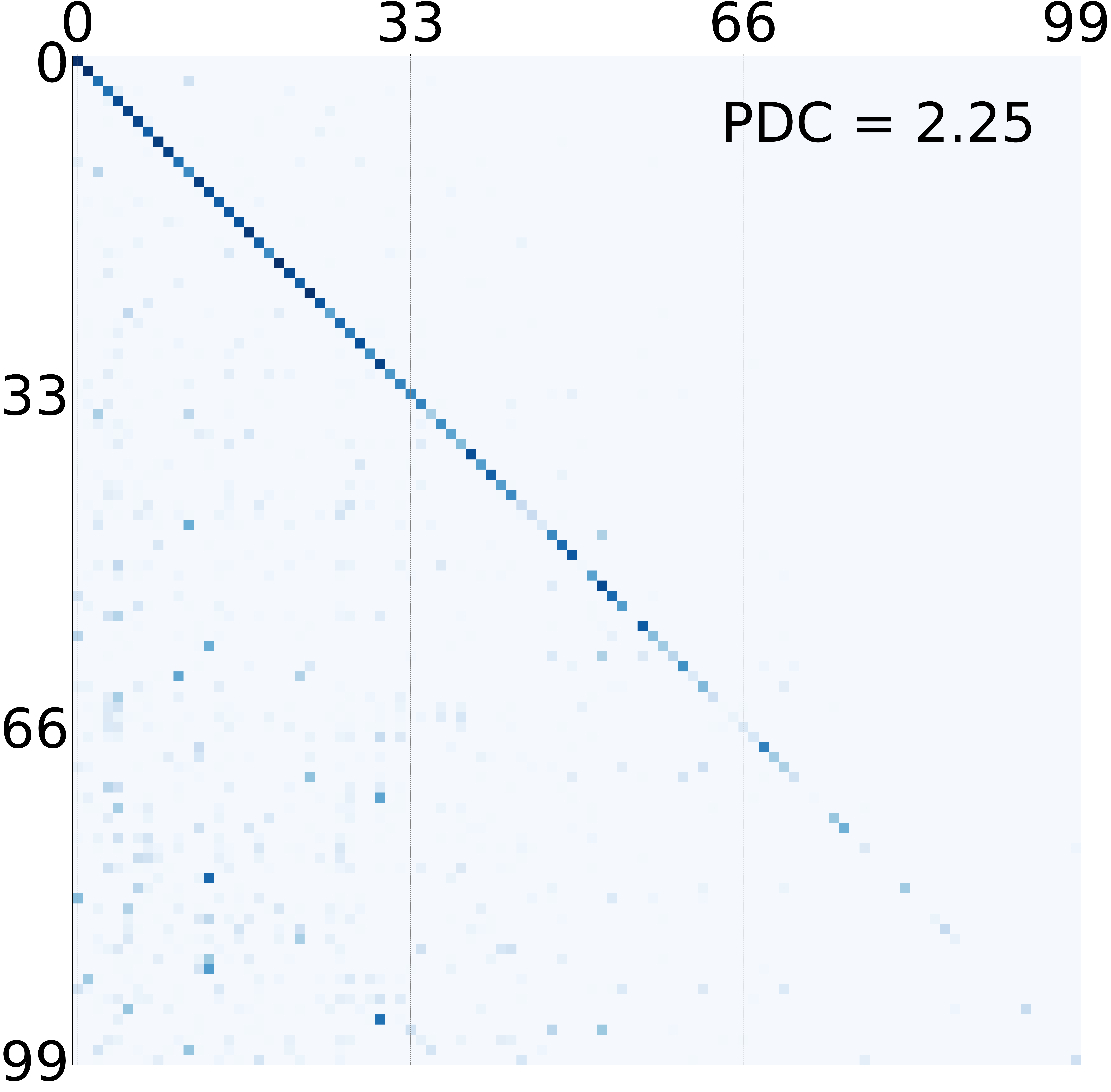} }
    \hspace{30pt}	
    \subfigure[MiSLAS]{
    \includegraphics[height=0.22\linewidth]{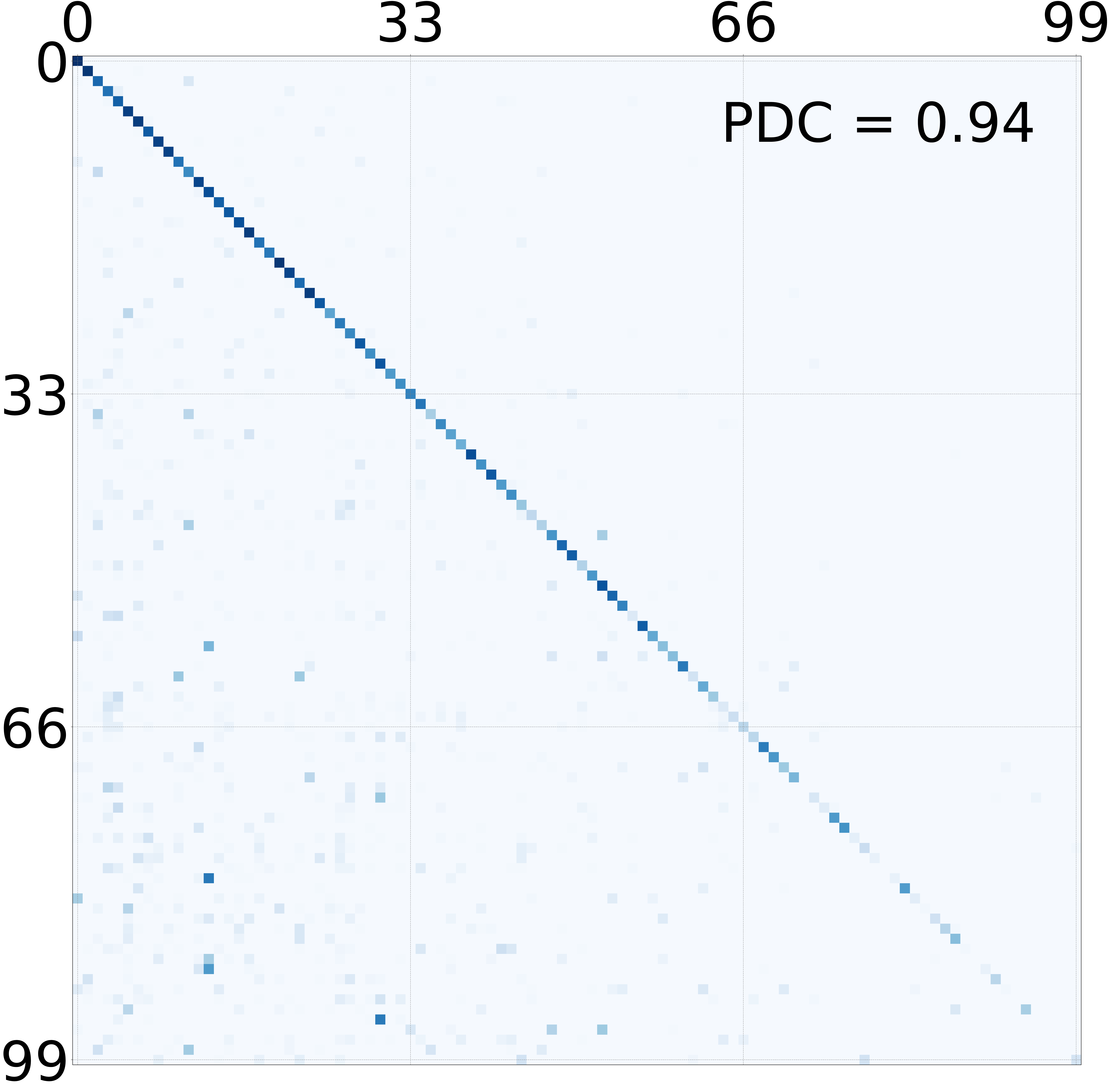} }
    \hspace{30pt}
    \subfigure[IB]{
    \includegraphics[height=0.22\linewidth]{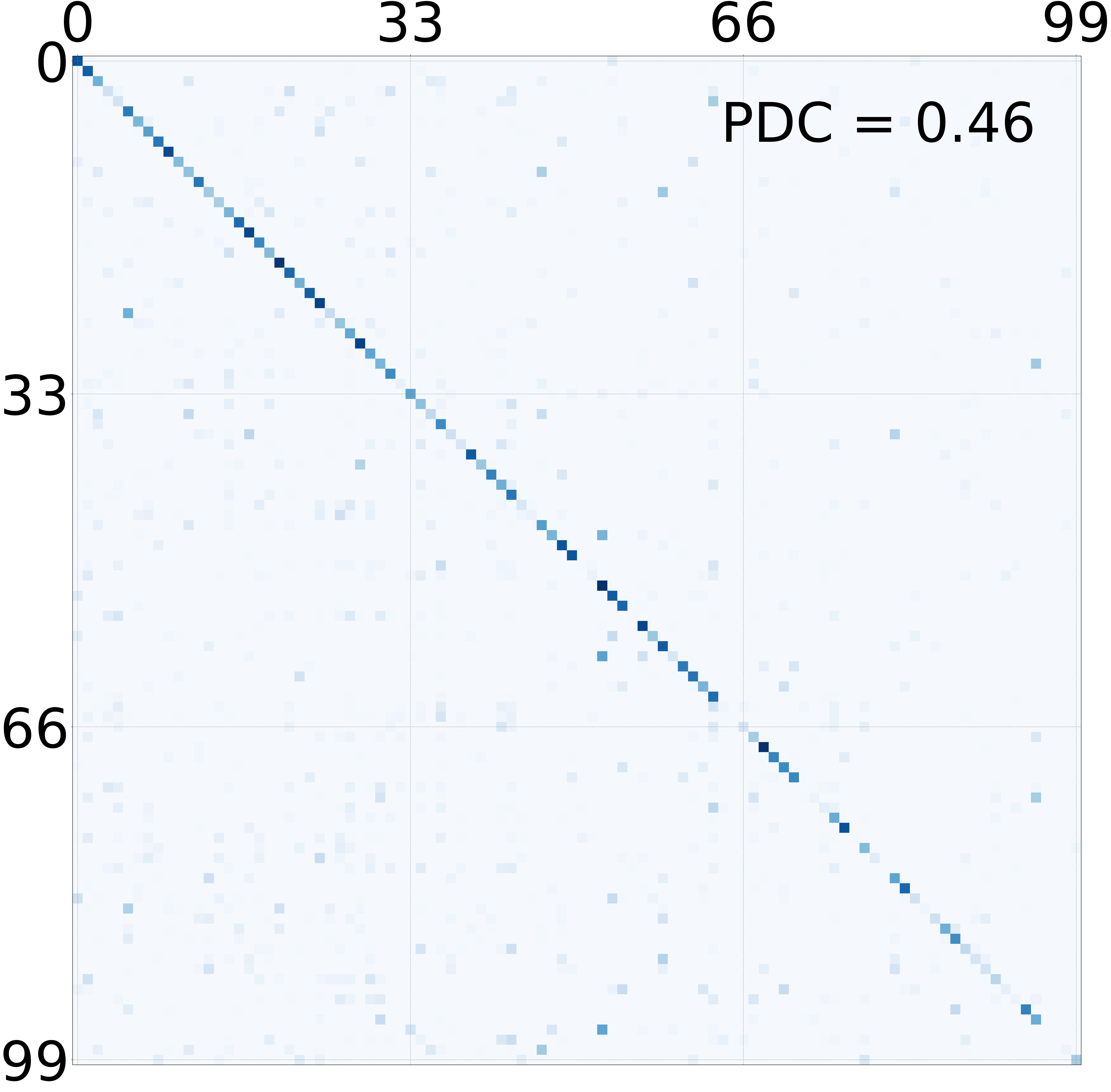} }
    \hspace{30pt}	
    \subfigure[BalCE]{
    \includegraphics[height=0.22\linewidth]{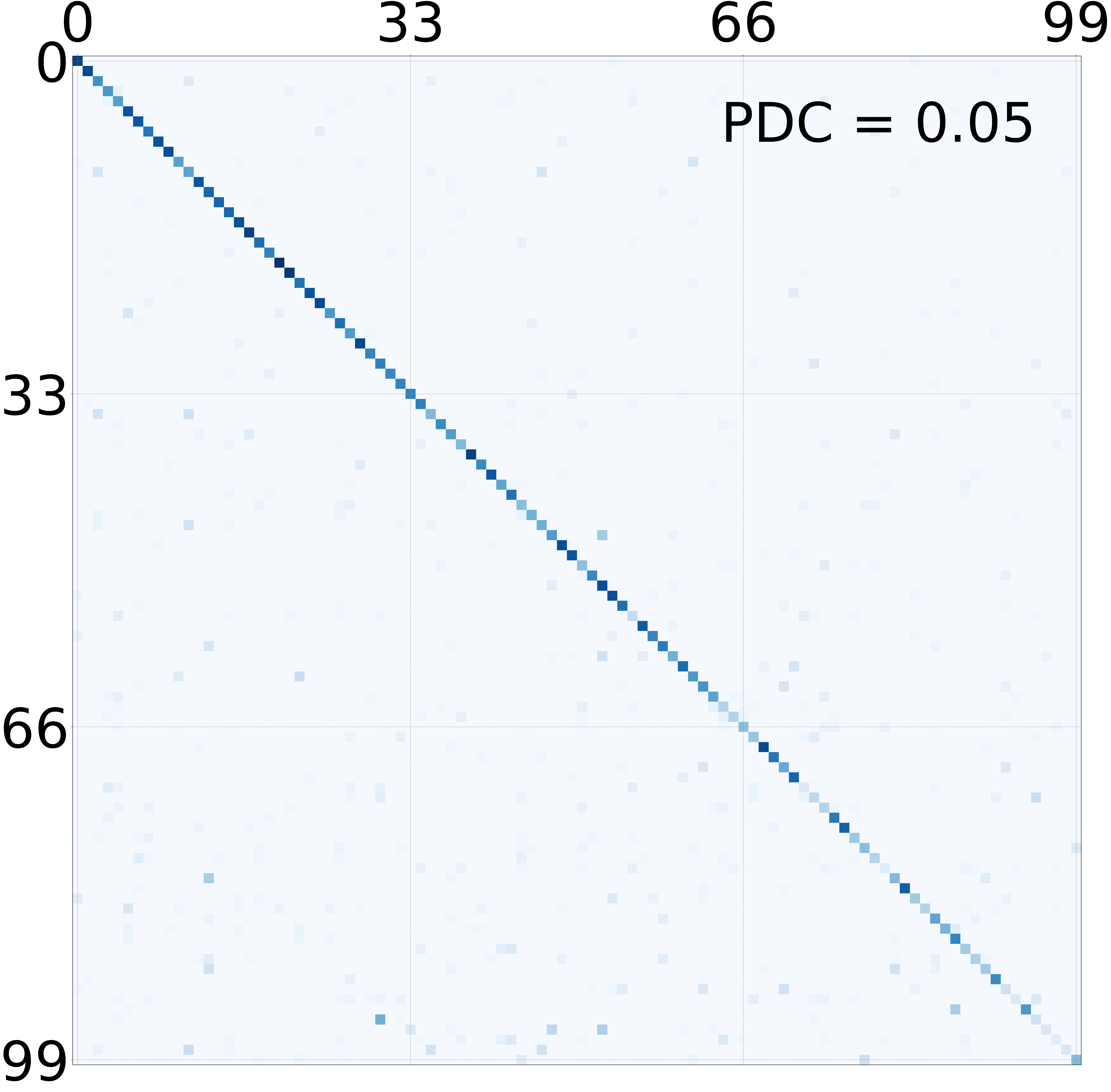}}
}
\vspace{-10pt}
\caption{Confusion matrix for each model on CIFAR-100-LT (IF=100). x-axis: predicted label. y-axis: ground truth.}
\label{fig:cf}
\vspace{-10pt}
\end{figure*}

%% file: Tabs/iNat18.tex
\begin{table}[t!]
\centering
\caption{ViT-B Performance on iNaturalist 2018. Bold indicates the best. $\dagger$: 224 resolution. *:ResNet50 performance.}
\resizebox{1\linewidth}{67pt}{
\begin{tabular}{l|cccc|c|c}
\hline
\multicolumn{1}{c|}{Method} & Many  & Med.  & Few   & Acc   & Acc*     & PDC  \\ \hline
CE                         & 72.45 & 62.16  & 56.62 & 61.03 & 57.30    & 0.76 \\
BCE\cite{Deit}             & \textbf{73.67} & 65.23  & 60.41 & 64.19 & 59.80    & 0.66 \\
CB\cite{CB-loss}           & 55.90 & 62.77  & 59.07 & 60.60 & 61.12    & 0.42 \\
LDAM\cite{LDAM-loss}       & 72.61 & 67.29  & 63.78 & 66.45 & 64.58    & 0.49 \\
MiSLAS\cite{MiSLAS}        & 72.53 & 64.70  & 60.45 & 63.83 & \textbf{71.60}    & 0.64 \\
LADE\cite{LADE-loss}       & 64.77 & 63.49  & 62.20 & 63.11 & 70.00    & 0.39 \\
IB\cite{IB-loss}           & 54.51 & 61.91  & 60.75 & 60.69 & 65.39    & 0.35 \\ 
BalCE\cite{BalCE}          & 67.82 & \textbf{68.36}  & \textbf{67.34} & \textbf{67.90} & 69.80    & \textbf{0.27} \\ \hline
CE$\dagger$                & 81.35 & 72.37  & 67.45 & 71.35 & -        & 0.50 \\
BCE\cite{Deit}$\dagger$    & 82.54 & 74.85  & 70.42 & 73.89 & -        & 0.42 \\
BalCE\cite{BalCE}$\dagger$ & \textbf{77.83} &\textbf{ 77.73}  & \textbf{76.95} & \textbf{77.43} & -    & \textbf{0.18} \\ \hline
\end{tabular}}
\vspace{-10pt}
\label{tab:inat}
\end{table}


%% file: Tabs/Rebuttal.tex
\begin{table}[t]
\centering
\caption{Performance of ViT-B w/o pretrained weights (CIFAR) or MAE pretraining (iNat18, 128 resolution).}
\resizebox{1\linewidth}{40pt}{
\begin{tabular}{l|cccc|cc}
\hline
\multirow{3}{*}{Method} & \multicolumn{4}{c|}{CIFAR100-LT}                                 & \multicolumn{2}{c}{\multirow{2}{*}{iNat18}} \\ \cline{2-5}
                        & \multicolumn{2}{c|}{IF=10}         & \multicolumn{2}{c|}{IF=100} & \multicolumn{2}{c}{}                        \\ \cline{2-7} 
                        & Acc   & \multicolumn{1}{c|}{PDC}   & Acc           & PDC         & Acc                   & PDC                 \\ \hline
CE                      & 18.69 & \multicolumn{1}{c|}{12.20} & 11.34         & 6.74        & 39.02                 & 1.30                \\
BCE\cite{deit3}                     & 17.26 & \multicolumn{1}{c|}{13.76} & 9.86          & 7.24        & 42.01                 & 1.25                \\
MiSLAS\cite{MiSLAS}                  & 18.70 & \multicolumn{1}{c|}{12.40} & 11.40         & 9.06        & 40.23                 & 1.19                \\
BalCE\cite{BalCE}                   & 20.93 & \multicolumn{1}{c|}{1.95}  & 15.89         & 1.26        & 42.01                 & 0.43                \\ \hline
\end{tabular}}
\vspace{-10pt}
\label{Tab:apdx}
\end{table}


%% file: Texs/5.conclusion.tex
\section{Conclusion}
In this paper, we rethink the performance of LTR methods with Vision Transformers and propose a baseline based on unsupervised pre-train to learn imbalanced data. We re-analyze the reasons for LTR methods' performance variation based on ViT backbone. Furthermore, we propose the PDC to measure the model predictive bias quantitatively, i.e., the predictors prefer to classify images into common classes. Extensive experiments demonstrate the effectiveness of PDC, which provides consistent and more intuitive evaluation.